\pdfoutput=1

\documentclass[11pt]{article}

\usepackage{acl}

\usepackage{times}
\usepackage{latexsym}

\usepackage[T1]{fontenc}

\usepackage[utf8]{inputenc}

\usepackage{microtype}
\usepackage{colortbl}
\usepackage{times}
\usepackage{latexsym}
\usepackage{amsmath}
\usepackage{amssymb}
\usepackage{times}
\usepackage{latexsym}
\usepackage{xcolor}
\usepackage{bm}
\usepackage{multirow}
\usepackage{booktabs}
\usepackage{graphicx}
\usepackage{pifont}
\usepackage{tikz}
\usepackage{soul}
\usetikzlibrary{bayesnet,arrows,positioning,calc}
\DeclareMathOperator*{\argmaxB}{arg\ max}

\DeclareMathOperator*{\argtopk}{arg\ topk}
\newcommand{\eba}[0]{\textsc{HUG}}
\newcommand{\fulleba}[0]{\textsc{Hop, Union, Generate}}
\newcommand{\bridge}[1]{{\color{red} \textbf{\textit{#1}}}}
\newcommand{\support}[1]{{\color{blue}\textit{#1}}}

\newcommand{\entity}[1]{{\color{purple}\textbf{#1}}}

\usepackage{arydshln}

\title{\fulleba: \\ Explainable Multi-hop Reasoning without Rationale Supervision}

\author{Wenting Zhao \and Justin T. Chiu \and Claire Cardie  \and Alexander M. Rush\\
        Department of Computer Science \\ Cornell University \\ \texttt{\{wz346,jtc257,ctc9,arush\}@cornell.edu}}

\begin{document}
\maketitle
\begin{abstract}
    Explainable multi-hop question answering (QA) not only predicts answers but also identifies rationales, i. e. subsets of input sentences used to derive the answers.
    This problem has been extensively studied under the supervised setting, where both answer and rationale annotations are given.
    Because rationale annotations are expensive to collect and not always available, recent efforts have been devoted to developing methods that do not rely on supervision for rationales.
    However, such methods have limited capacities in modeling interactions between sentences, let alone reasoning across multiple documents.
    This work proposes a principled, probabilistic approach for training explainable multi-hop QA systems without rationale supervision.
    Our approach performs multi-hop reasoning by explicitly modeling rationales as sets,
    enabling the model to capture interactions between documents and sentences within a document.
    Experimental results show that our approach is more accurate at selecting rationales than the previous methods, while maintaining similar accuracy in predicting answers.
\end{abstract}
\section{Introduction}
Multi-hop reasoning is an important capability for any intelligent machine comprehension system.
Question answering (QA) is a common application for evaluating a system's ability to reason across multiple steps~\cite{geva-etal-2021-aristotle,yang-etal-2018-hotpotqa,welbl-etal-2018-constructing}.
Large language models have achieved tremendous success on challenging QA tasks, even in the few-shot setting~\cite{wei2022chain}.
However, \citet{min-etal-2019-compositional} and \citet{chen-durrett-2019-understanding} demonstrate that these models, in reality, often bypass multi-hop reasoning by performing shallow pattern matching, resulting in poor generalization ability~\cite{tang-etal-2021-multi}. 
To avoid predictions made from such reasoning shortcuts, it is important to understand the series of steps the systems follow to derive the answers.

\begin{figure}
    \centering
    \includegraphics[width=\linewidth]{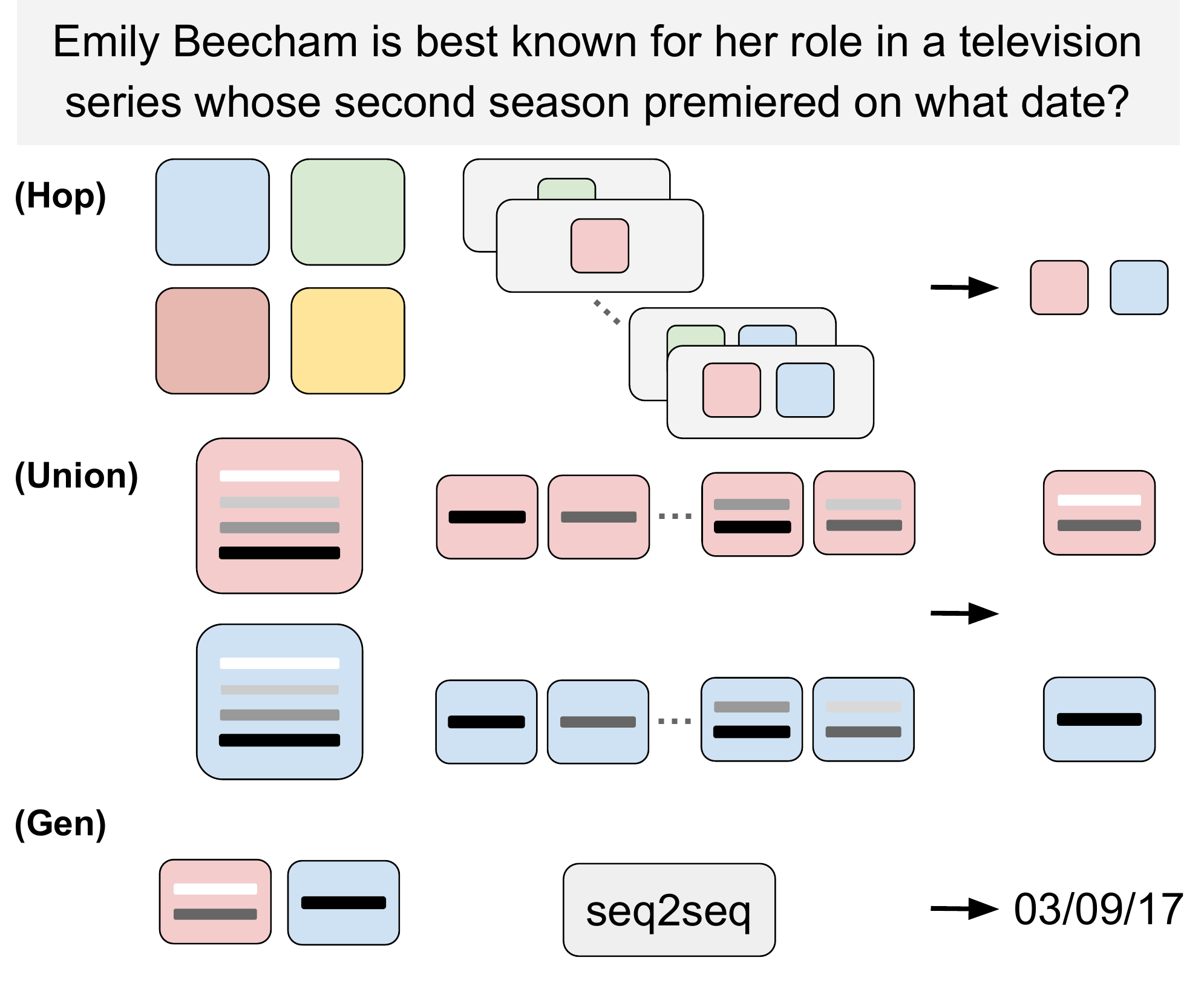}
    \caption{
    An overview of \eba, which proceeds in three stages. \textbf{Hop} explicitly considers all possible document sets and selects the most likely document set, \textbf{Union} explicitly considers all sentence subsets and chooses the most likely sentence subset within each selected document, and \textbf{Generate} combines the chosen sentence subsets and generates an answer.
    }
    \label{fig:overview}
\end{figure}

This work explores the challenge of building explainable multi-hop QA systems, which, in addition to predicting an answer, also identify a \emph{rationale} -- the set of sentences that lead to the answer.
Depending on task specifications, the rationale can be within a single document, or span across multiple documents.
Explainable multi-hop QA has been extensively studied in the supervised setting, where both rationale annotations and answer annotations are given.
These approaches either apply multi-task loss functions~\cite{joshi-etal-2020-spanbert,groeneveld-etal-2020-simple,deyoung-etal-2020-eraser} or design specialized network architectures~\cite{tu-etal-2019-multi,fang-etal-2020-hierarchical}.
However, having access to rationale annotations is a strong assumption.
In practice, they are expensive to collect~\cite{geva-etal-2021-aristotle}, less available than answer annotations~\cite{welbl-etal-2018-constructing}, and can suffer from low agreement rates between annotators~\cite{zhang-etal-2020-winowhy}.
Researchers have thus explored approaches that do not require rationale annotations~\citep{NEURIPS2020_6b493230,glockner-etal-2020-think,atanasova2022diagnostics}.
However, these previous approaches limit their reasoning to information from 1 or 2 sentences, and so they cannot be applied in \textit{multi-hop} scenarios, i.e. QA tasks that require making connections between several pieces of information across sentences and across documents. Additionally, these methods are either restricted to only work for multiple-choice QA, or restricted to only produce rationales at the document level but not at the sentence level.
We propose \fulleba \ (\eba), a principled, probabilistic approach for training explainable multi-hop QA systems without rationale supervision.
\eba \ overcomes the two-sentence limitation of previous methods by directly reasoning about rationales as \textit{sets} of sentences, while also extending rationale prediction to the multi-document setting.
We show an overview of \eba \ in Figure~\ref{fig:overview}.
\eba \ leverages the naturally hierarchical structure of text and proceeds in three stages -- it first selects the relevant set of documents given the question (Hop); then, it selects a subset of sentences within each of the relevant documents and collects them together (Union); finally, it generates an answer via a seq2seq model with all the collected sentences (Generate).
The key to multi-hop reasoning in \eba \ is modeling each selection as an explicit distribution over sets.
A probabilistic set distribution compares non-contiguous and variable size rationales, affording \eba \ flexibility for rationale selection.

Training a set-prediction model quickly becomes intractable as the size increases.
We make two algorithmic choices that lead to tractable training for \eba.
Treating rationales as a latent variable requires \eba \ to marginalize over all possible rationales, leading to an intractable learning objective.
\eba \ overcomes this issue by performing sampling in a hierarchical way -- it first identifies the most promising documents and then the most promising sentences within those documents.
Second, multi-hop QA often involves reasoning over long documents,
which is challenging due to the computational complexity of encoding long documents with neural models such as transformers. 
To make this encoding efficient, \eba \ performs computation in the embedding space.


We empirically evaluate \eba \ on three different multi-hop QA datasets: HotpotQA~\cite{yang-etal-2018-hotpotqa}, MuSiQue~\cite{trivedi2021musique}, FEVER~\cite{deyoung-etal-2020-eraser}, and MultiRC~\cite{deyoung-etal-2020-eraser}.
Results show that, for both selecting rationales and predicting answers, \eba \ is better than a number of state-of-the-art semi-supervised and unsupervised methods~\cite{chen2019multi,NEURIPS2020_6b493230,glockner-etal-2020-think,atanasova2022diagnostics} on all of the datasets.
We also demonstrate that \eba \ combined with larger language models consistently produces better performance.
We then analyze performance according to different types of multi-hop reasoning and show that by explicitly modeling multi-hop reasoning, \eba \ achieves a large improvement on the reasoning type that requires bridging entities.
\section{Related Work}
\paragraph{Explainable methods for multi-hop QA.}
Active research has been devoted to collecting human rationales for a wide range of QA tasks; a recent survey has identified 65 datasets that provide explanation annotations~\cite{wiegreffe2021teach}.
The appearance of such datasets has enabled rapid progress in supervised methods for extracting reasoning chains; we refer readers to~\citet{thayaparan2020survey} for a comprehensive survey.
While these supervised methods such as \citet{qi-etal-2019-answering} have achieved tremendous success on retrieving rationales, even in the open-domain setting~\footnote{While we only consider the distractor QA setting in this work, \eba \ can be combined with a rule-based retrieval system such as BM25 to be adapted to the open-domain setting.}, they can only be applied when rationale annotations are available.
However, such annotations do not always exist -- \citet{welbl-etal-2018-constructing} and \citet{yu2022crepe} propose two complex reasoning datasets that do not have rationale annotations.
In these cases, we need unsupervised rationale selection methods to still build explainable QA systems.



Other works have explored multi-hop QA with only answer supervision but not rationale supervision. As in our work, Retrieve and Generate (\textsc{RAG}) \cite{NEURIPS2020_6b493230}, treats the rationale as a latent variable;
however, in \textsc{RAG} the open retrieval-stage is to find a single document, ignoring the connections between different documents. \textsc{RAG} also does not produce sentence-level rationales. Other works consider sentence rationales:
\citet{glockner-etal-2020-think} compute a score for every sentence pair and pick the sentence pair that has the highest score as the rationale for answer prediction, and
\citet{atanasova2022diagnostics} perform binary classification on whether individual sentences are included in the rationale with added constraints such as consistency and faithfulness.
The shared limitation of these methods is that they do not capture the dependency between more than two pieces of information.
\eba \ overcomes this limitation by performing multi-hop reasoning as document set prediction and sentence set prediction.




Outside of unsupervised methods,
\citet{chen2019multi} propose a semi-supervised method which collects silver rationale annotations.
However, their method is limited to bridge-based questions, which is only one form of multi-hop reasoning; the other types can be found in \citet{trivedi2021musique}.
\vspace*{-5pt}
\paragraph{Rationales as latent variables.}
A focus for rationale methods in NLP outside of multi-hop QA has been identifying subsets of input tokens to justify decisions.
For text classification, \citet{lei-etal-2016-rationalizing}, \citet{bastings-etal-2019-interpretable}, and \citet{chen-ji-2020-learning} frame rationales as minimal subsets of input tokens.
For multi-hop QA, where input tokens are too granular a representation for rationales, treating sentences as rationales within long documents leads to the challenges of hierarchical selection and the representation of long documents; both of which we address with \eba.

Outside of using input tokens for rationales, \citet{zhou2020towards} assume rationales take the form of unconstrained text. While flexible, this approach leads to computationally expensive training methods.
Therefore, we constrain rationales to be a set of sentences from given documents, which both accommodates the production of useful intermediate reasoning steps and keeps training tractable. 
\vspace*{-5pt}
\paragraph{Unsupervised evidence retrieval.}
A task closely related to our setting (i.e., no access to rationale supervision) is unsupervised evidence retrieval, which searches for sentences relevant to the questions but does not predict answers.
For example, one could apply \citet{yadav-etal-2019-quick}, \citet{yadav-etal-2020-unsupervised}, \citet{zhao-etal-2021-distantly} and \citet{xu-etal-2021-exploiting-reasoning} to first identify the rationale for a multi-hop QA example and predict an answer based only on the rationale rather than on the entire document, so that the answer prediction is more constrained.
\vspace*{-5pt}
\section{Generative Multi-Hop QA}
\begin{figure}
    \centering
    \small
    \framebox{
    \parbox{0.45\textwidth}{
    \small
    $x$: Emily Beecham is best known for her role in a television series whose second season premiered on what date? \noindent \smallskip \newline
    $d_1$: \textbf{[1]} \support{\underline{Emily Beecham} is an English-American actress.} \textbf{[2]} \support{\underline{She} is best known for her role in the AMC television series ``\bridge{Into the Badlands}''}  \textbf{[3]} In 2011, she received the Best Actress award at the London Independent Film Festival.\smallskip \newline
    $d_2$: \textbf{[4]} \support{\underline{\bridge{Into the Badlands}} is an American television series that premiered on AMC November 15, 2015} \textbf{[5]} The series features a story about a warrior and a young boy who journey through a dangerous feudal land together seeking enlightenment. \textbf{[6]} \support{AMC renewed \underline{the show} for a 10-episode second season, which premiered on March 19, 2017.} \textbf{[7]} On April 25, 2017, AMC renewed the series for a 16-episode third season.'\smallskip \newline
    $\bm{z}$: [1], [2], [4], [6] \smallskip \newline
    $y$: March 19, 2017}
    }
    \caption{A QA example. The rationale $\bm{z}$ used to derive the answer is highlighted in \support{blue italics}, the document-level interaction is highlighted in \bridge{red boldface}, and the sentence-level interaction (i.e., coreference resolution) is highlighted in \underline{underline}. \eba \ models dependencies both between documents and between sentences within a document, thus being equipped with the capacity to perform multi-hop reasoning.}
    \label{fig:multihop_example}
\end{figure}

In the standard multi-hop QA setting, an example consists of a question $x$, a set of documents $D$, and an answer $y$.
Within $D$, some documents are relevant to the question, while the others are distractors. 
Explainable multi-hop QA models predict a rationale $\bm{z}$, a minimal set of sentences across the relevant documents, in addition to predicting the answer $y$.
We show a multi-hop QA example (with distracting documents omitted) in Figure~\ref{fig:multihop_example}.


\subsection{Model}

We propose the following generative model for multi-hop QA. Given the question $x$, we first select a subset of documents $\bm{d}=\{d_1,d_2,\ldots\}\subseteq D$. Next, within each document $d_i$, we select a subset of sentences $\bm{z}_i$. Finally, conditioned on the union of sentence sets from each document, $\bm{z}=\cup_i\bm{z}_i$, we generate an answer $y$. The only assumption we make in the model is that sentence sets are selected independently among documents.
Formally, we write the model as,
\begin{align}
    p(\bm{d}, \bm{z}, y \mid x) &= p(\bm{d} \mid x) \label{doc} \\
    &\quad \cdot \prod_{i} p(\bm{z}_i \mid d_i, x) \label{rationale} \\
    &\quad \cdot p(y\mid\bm{z}, x). \label{answer}
\end{align}
We refer to Eq.~\ref{doc} as the document set selection model, Eq.~\ref{rationale} as the sentence set selection model,
Eq.~\ref{answer} as the answer generation model.

\paragraph{Document Set Selection}
We select a set of documents $\bm{d}$ by directly parameterizing a distribution over all valid document sets.
We rely on a document set scoring function $f(\bm{d}, x)$, which captures both the relevance of
the document set $\bm{d}$ to the question $x$, as well as the dependencies among the documents in the set.
The document set selection model is given by
\begin{equation*}
    p(\bm{d} \mid x) \propto \exp(f(\bm{d}, x)).
\end{equation*}
This distribution is globally normalized over all valid subsets of documents $D$,
requiring the evaluation of the document scoring function $f$ on all valid document subsets.
Document set validity is dataset specific, and is discussed in Section \ref{sec:experiments}.

For efficiency, the document set scoring function $f$ first computes embeddings of each document in the set $\bm{d}$ independently, then combines them with a neural network (MLP).
Formally, let $\text{emb}: \mathcal{V}^* \rightarrow \mathbb{R}^n$ be an embedding function that maps a sequence of text to an $n$-dimension vector, where $\mathcal{V}$ is the vocabulary.
The document set scoring function is given by
\begin{equation*}
    f(\bm{d}, x) = \text{MLP}(\text{emb}(d_1,x), \text{emb}(d_2,x), \ldots).
\end{equation*}
We provide the details of the MLP in Appendix \ref{sec:document-scorer} and the details of the embedding function below, as part of the sentence selection model description.


\paragraph{Sentence Set Selection}
Within each document $d_i$, we select $\bm{z}_i \in \mathcal{P}(d_i)$, a power set of all sentences in $d_i$.
We rely on a sentence set scoring function $g(\bm{z}_i, x)$, similar to the document set scoring function,
which captures all relationships between selected sentences and the question.
The sentence set selection model is given by
\begin{equation*}
    p(\bm{z}_i\mid d_i, x) \propto \text{exp}(g(\bm{z}_i,x)),
\end{equation*}
which is globally normalized over all valid subsets of sentences in the document $d_i$.
Computing $p(\bm{z}_i\mid d_i, x)$ requires enumerating all sentences subsets, which is intractable.
We instead extend the approach of \citet{li2022easy}, which obtains document and contextual sentence representations in a single encoding step.
We insert a special \texttt{[SPC]} token at the beginning of each sentence, shown here at positions $k_1, k_2, \dotsc$,
\begin{align*} u = \texttt{[C}&\texttt{LS]} \text{\{x\}} \texttt{[SEP]} &\texttt{[S}&\texttt{PC]}\ldots &\texttt{[S}&\texttt{PC]}\ldots. \\
    &0 && k_1 && k_2
\end{align*}
We then obtain sentence subset $\text{emb}(\bm{z}_i)$ embeddings by feeding this to an encoder-only model such as BERT~\cite{devlin-etal-2019-bert} and taking the average of the contextual embeddings of the special tokens corresponding to the sentences in $\bm{z}_i$:
\begin{align*}
    \text{emb}(\bm{z}_i, x)&= \frac{1}{|\bm{z}_i|}\sum_{j \in \bm{z}_i}\text{encoder}(u)_{k_j}.
\end{align*}
Finally, let $v$ be a learnable vector, $g(\bm{z}_i, x) = v^T \text{emb}(\bm{z}_i, x)$
In practice, we note that encoder methods have a maximum input length, which can prevent full document encodings. We provide the details of long document encoding in Appendix~\ref{sec:encoding-long-docs}.
We also only consider subsets of up to a fixed max length.



\paragraph{Answer Generation}
Parameterization of the answer generation model, $p(y\mid\bm{z}, x)$, is done using a sequence-to-sequence model where the question and rational are fed to an encoder, and that answer is generated. This process is complicated by the fact that answers can take on different forms, depending on specific QA tasks such as Boolean QA, multiple-choice QA, extractive QA, and abstractive QA, etc.
We can therefore use a sequence-to-sequence model such as BART~\cite{lewis-etal-2020-bart} and T5~\cite{2020t5}, and provide prompt templates for the different variants of the task, given in the next section.


\subsection{Training and Inference}
To learn an explainable multi-hop QA system, \eba\ optimizes an approximation of the marginal likelihood. 
The marginal likelihood,
\begin{equation*}
    \mathcal{L(\theta)} = \text{log} \sum_{\bm{d}, \bm{z}} p_\theta(y, \bm{z}, \bm{d}\mid x),
\end{equation*}
is intractable, as it requires computing $p(y\mid\bm{z},x)$ under the answer generation model
for every valid set of sentences across documents.
We instead optimize a top-K Viterbi approximation of the marginal likelihood.
Given
\begin{align*}
    \mathcal{S}_{}^k &= \argtopk_{\bm{d}} p_\theta(\bm{d}\mid x)\\
    \mathcal{S}^k_{\bm{d}} &= \argtopk_{\bm{z}}
        \prod_i p_\theta(\bm{z}_i \mid d_i,x),
\end{align*}
we use the following approximation of the marginal likelihood
as our training objective:
\begin{equation*}
    \mathcal{L(\theta)} \approx \text{log} \sum_{\bm{d}\in \mathcal{S}^k} \sum_{\bm{z}\in \mathcal{S}^k_{d}} p_\theta(y, \bm{z}, \bm{d}\mid x).
\end{equation*}


At test time, we must choose the best documents and rationales.
Similar to training, we first choose the most likely pair of documents from $S^1$, then the most likely rationale from $S_{\bm{d}}^1$.
Finally, for span-based QA, we generate an answer by performing greedy search on the answer generation model $p(y|\bm{z},x)$; for Boolean QA or multiple-choice QA, we normalize the answers between different choices and take $\argmaxB_{y}p(y|\bm{z},x)$.

\section{Experimental Setup}
\label{sec:experiments}
\begin{table}[t]
\centering
\small
\begin{tabular}{p{0.07\linewidth} p{0.04\linewidth} p{0.75\linewidth}}
\toprule
& \cellcolor{blue!15} In          & \cellcolor{blue!15} {\noindent \color{red} \textit{A claim to be verified is that}} \{$x$\} {\noindent \color{red} \textit{We have following facts:}} \{$\bm{z}$\} \\
      & \cellcolor{blue!15} Out                        & \cellcolor{blue!15} {\noindent \color{red} \textit{The claim is thus}} \{supported/refuted\}. \\
BQA  & \cellcolor{green!15} In          & \cellcolor{green!15} {\noindent \color{red} \textit{A claim to be verified is that}} Steve Wozniak designed homes. {\noindent \color{red} \textit{We have following facts:}} Steve Wozniak primarily designed the 1977 Apple II , known as one of the first highly successful mass-produced microcomputers. \\

     & \cellcolor{green!15} Out                        & \cellcolor{green!15} {\noindent \color{red} \textit{The claim is thus}} refuted. \\ \midrule

& \cellcolor{blue!15} In         & \cellcolor{blue!15} {\noindent \color{red} \textit{Question:}} \{$x$\} \texttt{[SEP]} \{$\bm{z}$\} \\
           & \cellcolor{blue!15} Out                  & \cellcolor{blue!15} {\noindent \color{red} \textit{Answer:}} \{$y_1$\} (\{correct/wrong\}) \texttt{[SEP]} {\noindent \color{red} \textit{Answer:}} \{$y_2$\} (\{correct/wrong\}) \texttt{[SEP]} ...  \\
MCQ & \cellcolor{green!15} In         & \cellcolor{green!15} {\noindent \color{red} \textit{Question:}} Name few objects said to be in or on Allan 's desk. \texttt{[SEP]} Opening a side drawer, he took out a piece of paper and his inkpot. \\
           & \cellcolor{green!15} Out                  & \cellcolor{green!15}  {\noindent \color{red} \textit{Answer:}} Eraser  (wrong) \texttt{[SEP]} {\noindent \color{red} \textit{Answer:}} Inkpot (correct) \texttt{[SEP]} {\noindent \color{red} \textit{Answer:}} Pen (correct) \\ \midrule

 & \cellcolor{blue!15} In         & \cellcolor{blue!15} \{$x$\} \texttt{[SEP]} \{$\bm{z}$\} \\
           & \cellcolor{blue!15} Out                  & \cellcolor{blue!15} \{$y$\} \\
EQA & \cellcolor{green!15} In         & \cellcolor{green!15} Which American railroad, located in Southwestern Montana and Idaho, was backed by the Northern Pacific Railway? \texttt{[SEP]} The Gilmore and Pittsburgh Railroad (G\&P), now defunct, was an American railroad located in southwestern Montana and east-central Idaho. \\
           & \cellcolor{green!15} Out                  & \cellcolor{green!15} Gilmore and Pittsburgh Railroad \\
\bottomrule
\end{tabular}%
\caption{Seq-to-Seq Prompt templates for QA examples for FEVER as Boolean QA (BQA), MultiRC as multiple-choice QA (MCQ), and HotpotQA as extractive QA (EQA). Templates are in purple cells, followed by specific examples in green cells. Template keywords are highlighted in {\color{red} \textit{red italics}}.}
\vspace*{-10pt}
\label{tab:conversion}
\end{table}

\paragraph{Datasets and Their Representations.}
We evaluate \eba \ on four multi-hop QA datasets: HotpotQA in the distractor setting~\cite{yang-etal-2018-hotpotqa}, MuSiQue with answerable questions~\cite{trivedi2021musique}, FEVER~\cite{thorne-etal-2018-fever}, and MultiRC~\cite{khashabi-etal-2018-looking}.
HotpotQA and MuSiQue are extractive QA datasets that require reasoning over multiple Wikipedia documents and identifying a span of text as the answer.
In HotpotQA, each example contains ten candidate documents, and we must identify exactly two documents ($|D|=2$) that are relevant.
FEVER is a fact checking dataset that requires verifying claims made based on Wikipedia articles ($|D|=1$).
MultiRC is a multiple-choice QA dataset collected from diverse sources of documents including narrative stories and news articles, and their questions can only be answered by reasoning over multiple sentences ($|D|=1$).
Unlike conventional multi-choice tasks~\cite{lai-etal-2017-race,richardson-etal-2013-mctest}, MultiRC does not pre-specify the number of correct answer choices, resulting in a more challenging setting.
For FEVER and MultiRC, we consider the ERASER version~\cite{deyoung-etal-2020-eraser}, where rationale annotations are made cleaner and evaluation metrics are provided.

In Table~\ref{tab:conversion}, we demonstrate how to convert QA examples of these three datasets to a natural language prompt format~\cite{brown2020language}.
In FEVER, a claim $x$ needs to be classified as whether it is supported (1) or refuted (0) given the accompanying documents $D$.
For MultiRC, there is a varying number of choices per example, and an unknown number of the choices is correct; we stack all answer choices attached with their truth values as outputs to supervise the model.
Finally, extractive QA can naturally be formulated as a text-to-text problem, where the input is $x$ \texttt{[SEP]} $\bm{z}$, and the output is $y$.

\paragraph{Metrics and Comparison Systems.}
We compute F1 scores for rationale and document selection, and answer prediction.
F1 scores for rationales are computed at the sentence level.
Because the three QA datasets are in different formats, F1 scores for answers are computed differently.
For extractive QA, F1 scores are computed at the token level for the answer spans.
For Boolean QA and multiple-choice QA, F1 scores measure categorical answers.

For each dataset, we compare to (1) state-of-the-art approaches that require no rationale supervision and (2) at least one fully supervised method (i.e., answers and rationales available for training). The latter provides an upper bound on performance.\\
\noindent
\textbf{-- On HotpotQA and MuSiQue}, we compare to a rule-based approaches, BM25, and $\textsc{RAG}$~\cite{NEURIPS2020_6b493230} as unsupervised baselines.
We note that $\textsc{RAG}$ only performs document-level retrieval, and therefore its current form cannot be directly applied to identifying sentence-level rationales.
We modify $\textsc{RAG}$ to treat a sentence as a document, and at inference we take the top-3 sentences to be the rationale as it results in the highest sentence F1 scores.
For fair comparison, we parameterize $\textsc{RAG}$ in the same way as we parameterize \eba.
We also consider a semi-supervised approach -- \textsc{Chain}~\cite{chen2019multi}; they assume no access to gold rationale annotations and supervise their model on silver rationales produced with external entity taggers.
The fully supervised system we consider is SAE~\cite{tu2020select}.
Both \textsc{Chain} and SAE use RoBERTa-large as sentence encoders.\\
\noindent
\textbf{-- On FEVER and MultiRC}, we also compare to $\textsc{RAG}$ as an unsupervised baseline (predicting top-2 sentences for MultiRC, and top-1 sentence for FEVER).
Additionally, we consider diagnostics-guided explanation generation (\textsc{Diagnostics}) \cite{atanasova2022diagnostics} and faithful rationales (\textsc{Faithful}) \cite{glockner-etal-2020-think}.
Both of these methods have two variants -- one trained with rationale supervision (denoted by $\mathbb{RS}$-*) and the other trained without rationale annotations (denoted by $\mathbb{RU}$-*).
On MultiRC, We also compare to WT5~\cite{DBLP:journals/corr/abs-2004-14546}.

\paragraph{Implementation and Hyperparameters.}
We test \eba \ with language models of both small and large sizes.
For the small version (\eba-Small), we use distilBERT~\cite{Sanh2019DistilBERTAD} as the encoder and BART-base as the seq2seq model.
For the large version (\eba), we use RoBERTa-large as the encoder and BART-large as the seq2seq model.
We only test \textsc{RAG} in the small version.

We implement \eba \ with Hugging Face Transformers~\cite{wolf-etal-2020-transformers}.
We perform grid search with learning rates \{5e-6, 1e-5, 2e-5\} and batch sizes \{2,4,8,16\} for both \eba \ and \textsc{RAG}.
We train our system for 3 epochs for HotpotQA and 5 epochs for the other three datasets.
We warm up the learning rate with first 10\% examples.
We choose the checkpoint that has the highest answer F1 score on the validation set.
We consider rationales up to four sentences for HotpotQA and rationales up to three sentences for the other datasets.
%
Finally, we take $S^{10}$ and $S_{\bm{d}}^9$ for HotpotQA and $S_{\bm{d}}^{80}$ for MultiRC.
Because we remove rationales whose sentences are not contiguous in FEVER, we are able to compute the exact likelihood without top-k sampling.
\begin{table*}[t]
\centering

\begin{tabular}{lccccc}
\toprule
 & & \multicolumn{2}{c}{HotpotQA} & \multicolumn{2}{c}{MuSiQue} \\ 
 &\#Params& \multicolumn{1}{c}{\textbf{Sent F1}} & \multicolumn{1}{c}{Ans F1} & \multicolumn{1}{c}{\textbf{Sent F1}} & \multicolumn{1}{c}{Ans F1} \\ \midrule
BM25 & - & 40.5 & - & 12.9 & -\\
\textsc{RAG}-Small &221M& 49.0 & 62.8& 32.0 & 24.2\\
\eba-Small &221M& 67.1 & 66.8 &34.2&25.1\\
\eba & 761M & \textbf{72.5} & 73.5 &\textbf{44.4}&39.1\\ \midrule
\textsc{Chain} (semi-supervised) & 355M & 64.5 & 66.0 & - & -\\ 
SAE (supervised) & 790M  & 87.4 & 80.8 & 75.2 & 52.3\\
\bottomrule
\end{tabular}%
\caption{Performance comparison on predicting rationales and answers on HotpotQA and MuSiQue.}
\label{tab:hotpotqa}
\end{table*}

\begin{table*}[t]
\centering
\begin{tabular}{lrcccc}
\toprule
 & & \multicolumn{2}{c}{FEVER} & \multicolumn{2}{c}{MultiRC} \\ 
 &\#Params& \multicolumn{1}{c}{\textbf{Sent F1}} & \multicolumn{1}{c}{Ans F1} & \multicolumn{1}{l}{\textbf{Sent F1}} & \multicolumn{1}{l}{Ans F1} \\ \midrule
$\mathbb{RU}$-\textsc{Faithful} & 110M & 83.8 & 90.6 & 27.5 & 67.7 \\
$\mathbb{RU}$-\textsc{Diagnostics} & 110M & 56.1 & - & 38.1 & - \\
\textsc{RAG}-Small & 221M & 80.7 &92.0&44.9& 70.0\\
\eba-Small & 221M & 81.5 & 91.6 & 48.4 & 74.3 \\
\eba & 761M & \textbf{84.1} & 94.3 & \textbf{55.6} & 75.5 \\ \midrule
$\mathbb{RS}$-\textsc{Faithful} (supervised) & 110M & 91.4 & 91.1 & 46.1 & 67.4 \\
$\mathbb{RS}$-\textsc{Diagnostics} (supervised) & 110M & 94.4 & 89.7 & 79.4 & 71.7 \\
\bottomrule
\end{tabular}%
\caption{Performance comparison on predicting rationales and answers on Eraser-FEVER and Eraser-MultiRC. \eba \ has more parameters due to its use of a seq2seq model for answer generation.}
\label{tab:multirc_fever}
\end{table*}

\section{Results}
\label{sec:results}

\paragraph{HotpotQA.} We summarize the results on HotpotQA in Table~\ref{tab:hotpotqa}.
\eba-Small outperforms the best unsupervised approach \textsc{RAG}-Small by 18 sentence F1 points, demonstrating superior multi-hop reasoning abilities.
\eba-Small, despite having fewer parameters, outperforms the semi-supervised \textsc{Chain} on predicting rationales and is comparable to \textsc{Chain} on predicting answers.
While \textsc{Chain} explicitly exploits the heuristics used in the data collection process for HotpotQA, \eba-Small \ is able to learn such heuristics in a fully automatic way.
Finally, the gap between \eba \ and SAE, a fully supervised method that is given both rationale and answer annotations, remains large.

\paragraph{MuSiQue.} Table~\ref{tab:hotpotqa} shows that \eba-Small is better at both identifying rationales and predicting answers than \textsc{RAG}-Small, the best-performing unsupervised baseline.
Due to the difficulty of MuSiQue, it is harder for an unsupervised method to learn to select rationales -- the gap between the supervised method and the best unsupervised method on this dataset is greater than the gap on HotpotQA.

\paragraph{FEVER}
We summarize the results on FEVER in Table~\ref{tab:multirc_fever}.
In terms of selecting rationales in an unsupervised manner, \eba-Small outperforms $\mathbb{RU}$-\textsc{Diagnostics}, performs similarly to \textsc{RAG}-Small, and underperforms $\mathbb{RU}$-\textsc{Faithful} by a small margin.
Because FEVER mostly only requires single-hop reasoning, \eba-Small does not improve over the previous methods.
On predicting answers, \eba-Small outperforms $\mathbb{RU}$-\textsc{Faithful} but underperforms \textsc{RAG}-Small.
Compared to the supervised versions of $\mathbb{RS}$-\textsc{Diagnostics} and $\mathbb{RS}$-\textsc{Faithful}, which have access to both the answers and rationales during training, the gap between \eba-Small's and their rationale scores (sentence F1) understandably remains large. 



\paragraph{MultiRC}
Table~\ref{tab:multirc_fever} shows that \eba-Small outperforms all comparison models including \textsc{RAG}-Small -- the best competing approach without rationale supervision -- by 3.5 sentence F1 points and 4.3 answer F1 points.


\paragraph{Scaling \eba \ to Larger Models.}
On all three datasets, by increasing the number of model parameters, \eba \ can consistently achieve better performance. 
Additionally, as the number of reasoning hops increases, \eba \ can more benefit from the larger language models -- compared to \eba-Small, \eba \ has the least improvement on FEVER and had the most improvement on MuSiQue.
\begin{table}[t]
\centering
\small
\begin{tabular}{llcccc}
\toprule
\multicolumn{1}{l}{} & \multicolumn{1}{l}{} & \multicolumn{1}{c}{Sent F1} & \multicolumn{1}{c}{Doc F1} & \multicolumn{1}{c}{Ans F1} \\ \midrule
\multirow{2}{*}{Comparison} & \eba-Ind & \textbf{78.9} & \textbf{92.9} & 64.8 \\
 & \eba & 78.1 & 91.1 & \textbf{69.7} \\ \midrule
\multirow{2}{*}{Bridge} & \eba-Ind & 55.2 & 68.6 & 71.6 \\
 & \eba & \textbf{71.0} & \textbf{87.3} & \textbf{75.7} \\ \midrule
\multirow{2}{*}{Combined} & \eba-Ind & 60.0 & 73.4 & 69.1 \\
 & \eba & \textbf{72.5} & \textbf{88.0} & \textbf{73.5} \\
\bottomrule
\end{tabular}
\caption{Rationale selection performance broken down by different types of reasoning.}
\vspace*{-10pt}
\label{tab:breakdown}
\end{table}
\section{Analysis}
\paragraph{Document Dependencies}
\eba \ explicitly models the dependencies between documents for multihop reasoning.
We consider independent document selection~\footnote{We provide the detail of training and testing independent document selection model in Appendex~\ref{sec:ind-doc-select}.} to see whether this dependencies is necessary on the HotpotQA dataset.

To understand how document modeling impact rationale selection performance, we break the performance down by the reasoning types proposed in~\citet{yang-etal-2018-hotpotqa}: comparison-based reasoning and bridge-based reasoning.
In comparison-based reasoning, relevant documents independently contribute to the answer, whereas for bridge-based reasoning, relevant documents require connections to previously selected documents.
Table~\ref{tab:breakdown} summarizes answer F1 scores, document F1 scores, and sentence F1 scores.
While \eba-Ind is slightly better at comparison-based reasoning than the joint model, it fails at bridge-based reasoning; this result thus confirms the necessity of modeling the dependency between documents.
We also note that \eba-Ind and \eba \ have similar performance in predicting answers, but the gap between how accurate they select rationales is large, suggesting that \eba-Ind often derives answers with wrong reasoning.
Overall, \eba \ is better than \eba-Ind \ at both predicting answers and selecting rationales.

In addition to the quantitative analysis, we also qualitatively compare the two models in Figure~\ref{fig:qualitative}.
When considering paragraphs independently, documents A and C share the most entities with the question (i.e., Copsi, earl of Northumbria, and Two Rivers), so they are more likely to lead to the answer.
However, the correct documents are A and B. Deriving B not only depends on the question but also further requires knowing the information from A. Therefore, while having the independent document selection model can improve efficiency because it only performs one-step reasoning, the joint document selection model is necessary when reasoning steps depend on one another.

\begin{figure}[!t]
    \framebox{
    \parbox{0.45\textwidth}{
    \small
    \textbf{Q:} When \entity{Copsi} was made \entity{earl of Northumbria} he went to reside in a town at the confluence of which \entity{two rivers}?\noindent \smallskip \newline
    \textbf{Document A, Copsi:} \newline
    \entity{Copsi} survived Tostig's defeat at Stamford Bridge, and when William the Conqueror prevailed at Hastings he travelled, in March 1067, to pay William homage at Barking (where William was staying while his tower was being constructed in London). \support{In return, William made \entity{Copsi earl of Northumbria} and sent him back to York.}\newline
    \textbf{Document B, York:} \newline
    \support{York is a historic walled city at the confluence of the rivers Ouse and Foss in North Yorkshire, England.} The municipality is the traditional county town of the historic county of Yorkshire to which it gives its name.\newline
    \textbf{Document C, Two Rivers Press:} \newline
    \entity{Two Rivers} Press is an independent publishing house, based in the English town of Reading. Two Rivers Press was founded in 1994 by Peter Hay (1951–2003).\smallskip \newline
    \textbf{A:} Ouse and Foss}
    }
    \caption{A HotpotQA example where there is a dependency between two supporting documents, and thus selecting the second document independent of the first one results in insufficient information. Correct rationale is highlighted in \support{blue italics}. Entity overlaps between questions and documents are in \entity{red boldface}. \eba-Ind's predicted Documents B and C, whose reasoning remains at the surface level as they share the most entities with the question. \eba \ predicted Documents A and B, which demonstrates its ability of understanding dependency between documents.} \label{fig:qualitative}
\end{figure}
\begin{table}[t]
\centering
\small
\begin{tabular}{lcccc}
\toprule
Answer Model & Pred Type & \multicolumn{1}{l}{FEVER} & \multicolumn{1}{l}{MultiRC} \\
&& \multicolumn{2}{c}{\textbf{Sent F1}} \\
\midrule
BART & Generate &81.5 & 48.4 \\
RoBERTa & Classify & 82.0 & 20.0\\
\bottomrule
\end{tabular}
\caption{Comparison on sentence F1 scores between different parameterization choices of $P(y\mid\bm{z}, x)$.
}
\label{tab:answer_parameterization}
\end{table}

\begin{figure}[]
    \framebox{
    \parbox{0.45\textwidth}{
    \small
    \textbf{Q:} What did the judge tell Mr. Thorndike about the law? \noindent \smallskip \newline
    \textbf{A1:} {\color{blue}\textit{Cannot be swayed by wealth or political influences.}}\newline
    \textbf{A2:} {\color{blue}\textit{The law is not vindictive.}}\newline
    \textbf{A3:} {\color{blue}\textit{It was not vindictive.}}\newline
    \textbf{A4:} It was unjust.\newline
    \textbf{A5:} It was vindictive.\newline
    \textbf{A6:} {\support{The judge told Mr. Thorndike that the law is not vindictive. He said the law only wishes to be just. Judge said the law cannot be swayed by wealth or political influences.}}
    }
    }
    \caption{A test example from MultiRC that can be answered with commonsense reasoning and thus requires no accompanying documents. Correct answers are highlighted in \support{blue italics}.}
    \vspace*{-10pt}
    \label{fig:multirc_commonsense}
\end{figure}
\paragraph{Role of Answer Generation}
\eba \ uses a generative model (BART) to parameterize $p(y\mid \bm{z}, x)$. An alternative approach would be to use a classification model such as RoBERTa~\cite{DBLP:journals/corr/abs-1907-11692} to predict answers for FEVER and MultiRC (HotpotQA requires a generative model). 

Interestingly, Table~\ref{tab:answer_parameterization} shows the choice of answer model significantly impacts the ability of \eba \ to learn a rationale model. 
On FEVER, where claims cannot be verified without the corresponding rationales, BART and RoBERTa perform similarly. However, on MultiRC, where questions can often be answered without information in accompanying documents, the best Generative model outperforms the best Classification model by over 32 sentence F1 points.

Figure~\ref{fig:multirc_commonsense} shows an example of such a question where the answers can be guessed by the classification model using commonsense knowledge to reason about law.
Generative models need to assign a high probability to every token in the answer, and we hypothesize that they make better use of the answer supervision.

\paragraph{Speed evaluation.}
\begin{table}[t]
\centering
\small
\begin{tabular}{lccc}
\toprule
 & \eba & \textsc{Faithful} & Ratio \\ \midrule
Training & 11,544.2 & 444.0 & 26 \\
Inference &    39.0 & 3.9   & 10 \\
\bottomrule
\end{tabular}
\caption{Runtime comparison (in seconds). \eba \ uses 80 rationale samples at training time, and the argmax rationale at inference.}
\label{tab:overhead}
\end{table}

While \eba \ obtains strong sentence F1 scores,
training is more expensive because the model must consider a set of rationales for every example.
In particular, the answer model $p(y\mid \bm{z}, x)$ must be run for \emph{every} sampled $\bm{z}$ for each training example.
At inference, the answer model requires only a single evaluation of $p(y\mid \bm{z}, x)$ for $\argmaxB_{\bm{z}} p(\bm{z}\mid x)$.
We empirically measure the runtime overhead of \eba \ compared to \textsc{Faithful} on MultiRC, using 80 samples of $\bm{z}$ at training time.
We report the total training time and inference time in Table~\ref{tab:overhead}.
Compared to \textsc{Faithful}, \eba \ takes longer to train and to predict.
\section{Conclusion}
We present \eba, a probabilistic, principled approach for explainable multi-hop reasoning without rationale supervision.
\eba \ explicitly models multi-hop reasoning by considering the dependency between documents and between sentences within a document.
Experimental results demonstrate that \eba \ outperforms other state-of-the-art methods that do not rely on rationale labels.

\section*{Ethics Statement}
The goal of explainable methods is to improve the trustworthiness of systems.
\eba \ presents a method for fine-tuning language models for selecting rationales, without rationale annotations, that exploits the knowledge already present in pretrained language models.
While this has the potential of improving the trustworthiness of the model, it may also reinforce existing harmful biases in the language model.

\section*{Acknowledgement}

AR and JC are supported by a Sloan Fellowship, NSF CAREER \#2037519, and NSF \#1901030. CC and WZ are supported by NSF \#1815455.

\bibliography{anthology,custom}
\bibliographystyle{acl_natbib}

\appendix
\section{Document scoring function}
\label{sec:document-scorer}
Let $\text{MLP}: \mathbb{R}^{3n} \rightarrow \mathbb{R}$ be a multilayer perception:
$$\text{MLP}(x) = W_2\text{ReLU}(W_1x).$$
The document set scoring function is given by:
$$f(\bm{d},x)=\text{MLP}(\text{emb}(d_1,x),\ldots).$$
For document pairs, we specialize this to
\begin{align*}
f(\bm{d},x)&=\text{MLP}([\text{emb}(d_1, x),\text{emb}(d_2, x),s(d_1,d_2)])\\
s(d_1,d_2) &= |\text{emb}(d_1, x)-\text{emb}(d_2, x)|.
\end{align*}
For extending this parameterization to large document sets, we could use a similar parameterization to the sentence set scoring function:
$$f(\bm{d},x) = \text{MLP}(\sum_i \text{emb}(d_i,x)).$$

\section{Encoding long documents}
\label{sec:encoding-long-docs}
Transformer-based text encoders can only accept inputs shorter than a fixed length (e.g., 512 tokens).
To address this limitation, we partition documents into slices of $m$ sentences and compute the embedding for each slice individually.
We denote a slice for a document $d$ as $d^{i:j}$ that starts at the $i$th sentence and ends before the $j$th sentence.
Let $i \in \text{range}(0, |d|, p)$, we approximate $\text{emb}(d)$ by the following aggregation,
\begin{equation*}
    \text{emb}(d) \approx \left\lfloor \frac{|d|}{m}\right\rfloor \sum_i \text{emb}(d^{i:i+p}).
\end{equation*}
We set the slice length $m$ purely based on whether the longest slice is under 512 tokens.
$m$ is set to 3 for HotpotQA, 5 for FEVER, and 9 for MultiRC.

\section{Model selection}
\begin{table}[t]
\centering
\small
\begin{tabular}{lrr}
\toprule
Criteria & \multicolumn{1}{l}{Value} & \multicolumn{1}{l}{Sentence F1} \\
\midrule
Answer F1 & 66.81 & \textbf{67.05} \\
Answer EM & 53.51 & \textbf{67.05} \\
Answer NLL & 3.01 & 65.82 \\
\bottomrule
\end{tabular}
\caption{Sentence F1 scores from checkpoints that are chosen based on different criteria.}
\label{tab:model-selection}
\end{table}
In the unsupervised sentence selection setting, we cannot perform model selection by choosing the model with the highest validation sentence F1 score.
Instead, we must rely on answer evaluation measures:
validation answer F1, answer EM, or likelihood.
We train \eba \ for three epochs, checkpoint every 2500 steps,
and evaluate sentence F1 for the checkpoint with the best validation performance measure.
The results of these selection methods are presented in Table \ref{tab:model-selection}.
We find that performing model selection via both answer EM and answer F1 results in the best sentence F1, but the differences between different metrics are minor.

\section{Independent document selection model}
\label{sec:ind-doc-select}
For independent document selection, we train a different document selection model that factors as
\begin{equation*}
    p(\bm{d} \mid x) = \prod_{d\in \bm{d}} p(d \mid x).
\end{equation*}
Exact marginalization of $\bm{d}, \bm{z}$ is still intractable. We thus only marginalize over $\mathcal{S}^5$ and $\mathcal{S}_{\bm{d}}^5$.
At inference, we choose
\begin{equation*}
    \bm{d} = \argtopk_{d} p(d \mid x),
\end{equation*}
where $k$ is the number of documents pre-specified by the task.

\section{Revealing dataset shortcuts with \eba.}
\begin{figure}[!t]
    \framebox{
    \parbox{0.45\textwidth}{
    \small
    \textbf{Q:} Watertown International Airport and Blue Grass Airport, are in which country? \noindent \smallskip \newline
    \textbf{Document A, Blue Grass Airport:} \newline
    Blue Grass Airport is a public airport in Fayette County, Kentucky, 4 miles west of downtown Lexington.\newline
    \textbf{Document B, Watertown International Airport:} \newline
    Watertown International Airport is a county owned, public use airport located in Jefferson County, New York, United States.\smallskip \newline
    \textbf{A:} United States}
    }
    \framebox{
    \parbox{0.45\textwidth}{
    \small
    \textbf{Q:} Who is also an actor, Luis Llosa or Ron Howard? \noindent \smallskip \newline
    \textbf{Document A, Luis Llosa:} \newline
    Luis Llosa (born 1951) is a Peruvian film director.\newline
    \textbf{Document B, Ronald William Howard:} \newline
    Ronald William Howard (born March 1, 1954) is an American actor and filmmaker.\smallskip \newline
    \textbf{A:} Ronald William Howard}
    }
    \caption{Dataset Shortcuts. Two HotpotQA examples that do not need both documents to derive the answers.} \label{fig:insights}
\end{figure}

We show that \eba \ is able to discover examples in which answers can be derived with reasoning shortcuts.
\citet{yang-etal-2018-hotpotqa} claim that all HotpotQA examples require reasoning over two documents, but we identify a number of examples that fail this property with the following steps.
First, we look for the examples that \eba \ correctly predicts the answers but incorrectly predicts the rationales.
Of those examples, we look for a subset where only one document is correctly selected. 
Because if conditioning on the one document that can already lead to the answer, the other document is redundant.
Finally, we manually go through the filtered examples and find that many of the questions can be answered with one documents.
Figure~\ref{fig:insights} shows two types of reasoning shortcuts found by us.
The first question implies both airports are in the same country, and thus looking up one of the airports is sufficient.
In the second question, Document B alone contain the correct answer.
\end{document}